\documentclass[letterpaper]{article} 
\usepackage[preprint]{aaai2027}  
\usepackage[hyphens]{url}  
\usepackage{graphicx} 
\usepackage{natbib}  
\usepackage{caption} 
\usepackage{algorithm}

\usepackage{newfloat}
\usepackage{listings}
\DeclareCaptionStyle{ruled}{labelfont=normalfont,labelsep=colon,strut=off} 
\floatstyle{ruled}
\newfloat{listing}{tb}{lst}{}
\floatname{listing}{Listing}

\usepackage{booktabs}

\ifdefined\preamble\else
\usepackage{amsmath}
\usepackage{amssymb}
\usepackage{mathtools}
\usepackage{caption}
\usepackage{svg}
\usepackage{longtable}
\usepackage{array}
\usepackage{algorithmicx}
\usepackage{algcompatible} 
\usepackage{algpseudocode}
\usepackage{subcaption}
\usepackage{graphicx}
\graphicspath{{Figures/}}

\newcommand{\preamble}{}
\fi

\providecommand{\tightlist}{%
  \setlength{\itemsep}{0pt}\setlength{\parskip}{0pt}}

\newcommand*\phantomsection[1]{#1}
\newcommand*\href[1]{#1}
\makeatletter
\providecommand{\Gscale@div}{} 
\newsavebox\pandoc@box
\newcommand*\pandocbounded[1]{
  \sbox\pandoc@box{#1}%
  \Gscale@div\@tempa{\textheight}{\dimexpr\ht\pandoc@box+\dp\pandoc@box\relax}%
  \Gscale@div\@tempb{\linewidth}{\wd\pandoc@box}%
  \ifdim\@tempb\p@<\@tempa\p@\let\@tempa\@tempb\fi
  \ifdim\@tempa\p@<\p@\scalebox{\@tempa}{\usebox\pandoc@box}%
  \else\usebox{\pandoc@box}%
  \fi%
}
\makeatother
\def\fps@figure{htbp}

\title{Self-Repairing Recurrent Ensembles for \\Real-Time Recovery from Distribution Shift}
\author{
    Julian Lemmel\textsuperscript{\rm 1,\rm 2},
    Pedro D. Wendel Garcia\textsuperscript{\rm 2},
    Taisuke Kobayashi\textsuperscript{\rm 3},
    Radu Grosu\textsuperscript{\rm 1}
}
\affiliations{
    \textsuperscript{\rm 1}~Vienna University of Technology\\
    \textsuperscript{\rm 2}~Medical University of Vienna\\
    \textsuperscript{\rm 3}~National Institute of Informatics, Tokyo

    julian.lemmel@tuwien.ac.at
}

\begin{document}

\maketitle

\begin{abstract}
Deploying a pretrained controller exposes it to conditions that are absent from its training data. Sensor drift, outright sensor failure and accumulating measurement noise all induce a distribution shift that can collapse an otherwise competent policy – typically at a point in time where no expert is available to supply corrective labels. We present a method that lets a policy recover from such shifts online and without supervision. Our controller is an ensemble of recurrent networks, each of which observes a randomly masked subset of the observation vector, and whose Gaussian outputs are combined through sequential Kalman fusion so that confident members dominate the consensus action. At deployment, we treat this consensus as a self-supervised label and fine-tune each member towards it, scaling each member's contribution proportional to the complement of its squared Kalman gain. Gradients are computed using RFLO, an efficient and biologically plausible approximation of Real-Time Recurrent Learning, so that a parameter update follows every environment step and the policy reacts to a shift as it unfolds. On a range of simulated continuous control tasks, our approach recovers close to the original performance after a sensor shift, while ensembles that see the full observation are unable to recover. The same framework subsumes fully online interactive imitation learning: when an expert is present, the consensus label is replaced by the expert action and the identical update rule refines the policy during teleoperation.
\end{abstract}


\section{Introduction}\label{introduction}

A controller learned from data is only as good as the conditions it was trained under. Once it leaves the training distribution, its behavior is unconstrained -- and in a physical system, leaving the training distribution is not an edge case but an expectation. A joint encoder reads a constant offset after a hard impact or a failed homing procedure. A sensor stops reporting altogether and its channel is pinned to zero. Thermal drift and mechanical wear raise the noise floor gradually over hours of operation. In each case the policy continues to receive an observation vector of the expected shape, containing values it has never been trained on, and a controller that solved the task at deployment time can lose the ability to act entirely.

The established remedy is to bring an expert back into the loop. Interactive imitation learning (IIL), and the DAgger family of algorithms in particular \citep{ross2011}, addresses distribution shift by letting the novice act, having the expert label the states it visits, and retraining on the aggregated dataset. Safety-aware variants gate the novice's authority on a learned estimate of its own competence \citep{zhang2017a, kelly2019, menda2019}. These methods are effective, but they rest on an assumption that deployment rarely satisfies: that an expert is available at the moment the shift occurs. A robot that has been running unattended for a week has no expert on hand. Even when a human operator is present, providing a corrective label means specifying a continuous vector of joint commands -- something humans do poorly through a teleoperation interface, and cannot do at control-loop rates at all. Recovering from distribution shift, we argue, should not require supervision.

\begin{figure}[hb]
\centering
\pandocbounded{\includegraphics[width=0.7\linewidth, keepaspectratio,alt={Each member of the ensemble receives a different, randomly masked observation. If a sensor failure appears, some of the members don't experience it and are therefore not confused in their action prediction.}]{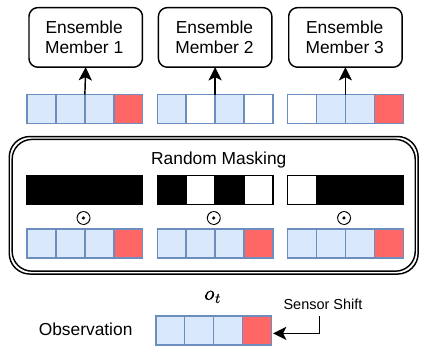}}
\caption{Each member of the ensemble receives a different, randomly masked observation. If a sensor failure appears, some of the members don't experience it and are therefore not confused in their action prediction.}\label{fig:ensemble}
\end{figure}

Our starting point is the observation that a sufficiently \emph{diverse} ensemble contains its own supervision. If each member of an ensemble is trained to solve the task from a different random subset of the observation dimensions, see Figure \ref{fig:ensemble}, then a shift that corrupts one dimension corrupts only those members that depend on it. The remaining members are unaffected and continue to predict sensible actions. What is needed is a way to identify them at runtime and to move the affected members back towards them.

We obtain both from the same mechanism. Each member is a stochastic policy predicting a Gaussian over actions, and we combine the members through sequential Kalman-gain fusion, following \citet{valencia2025}, so that the fused action is dominated by members that are confident and discounts members whose predicted variance is high. Members confused by a corrupted input are precisely the ones that report high variance, so the fused distribution is already the ensemble's best available estimate under the shift. We then close the loop: the mean of the fused distribution serves as a self-supervised label, and each member is fine-tuned towards it, weighted by the complement of its own contribution to the fusion. This is a follow-the-leader scheme in which the leader is the consensus, and it is related in spirit to the majority-voting objective \citet{zuo2025} use for test-time adaptation of language models, transposed from discrete votes to a continuous action distribution.

Applying this correction \emph{while the policy runs} is what makes it useful, and it is where the recurrent structure of the policy becomes a constraint. Our members are recurrent neural networks, which are necessary because the shifted observation must be interpreted in the context of the history that preceded it. Training recurrent networks conventionally means backpropagation through time (BPTT), which requires buffering a sequence, halting, and replaying it -- incompatible with a controller that must react within a control cycle. We therefore compute gradients with Random Feedback Local Online Learning (RFLO) \citep{murray2019}, a biologically plausible approximation of Real-Time Recurrent Learning (RTRL) \citep{williams1989} that maintains a Jacobian trace alongside the hidden state. A parameter update is available at every environment step, at no additional latency beyond the forward pass, and the policy adapts as the shift unfolds rather than after the episode has ended.

The resulting algorithm requires no expert and no separate training phase, and the same update rule serves both regimes: substituting an expert action for the consensus label recovers a fully online interactive imitation learning method, which we call RTR-IIL, and which we evaluate as well.

Our contributions are as follows.

\begin{itemize}
\tightlist
\item
  We introduce a recurrent ensemble policy with random observation masking, combined through Kalman-gain fusion, whose members are individually robust to the failure of any single observation dimension.
\item
  We propose a self-supervised follow-the-leader loss that uses the fused consensus action as a training label, enabling recovery from distribution shift with no expert supervision.
\item
  We show how to optimize this objective in real time using RFLO, yielding a controller that applies a parameter update at every environment step during operation.
\item
  We evaluate on simulated continuous control tasks under three distinct distribution shifts -- a shifted sensor reading, a failed sensor, and gradually increasing measurement noise -- and show that random observation masking is necessary for recovery: ensembles with full observation visibility fail to recover.
\end{itemize}

The remainder of the paper is structured as follows. Section 2 reviews interactive imitation learning and online gradient computation for recurrent networks. Section 3 introduces the ensemble architecture, the Kalman fusion rule and the self-supervised objective. Section 4 presents our experimental evaluation, and Section 5 concludes.

\section{Related Work}\label{related-work}

\textbf{Robustness to sensor corruption.} A common approach to sensor failure is to train for it explicitly, via domain randomization or input dropout, producing a fixed policy that tolerates a range of corruptions \citep{liu2017, tobin2017}. This yields robustness without adaptation, and is therefore limited to the corruptions anticipated at training time. Our random observation masking plays a similar role during pretraining, but we treat it as a means of manufacturing ensemble diversity rather than as the robustness mechanism itself -- the recovery comes from online fine-tuning, and we show in Section 4 that masking alone, without adaptation, does not restore performance.

\textbf{Ensemble policies and uncertainty.} Ensembles are widely used as a cheap approximation to a posterior over policies, and in imitation learning specifically to decide when a novice should defer to an expert \citep{menda2019, kobayashi_2026}. We use an ensemble for a different purpose: as a source of self-supervision. Our fusion rule follows \citet{valencia2025} in combining Gaussian predictions through sequential Kalman updates, which weights members by their own reported confidence. 

\textbf{Test-time adaptation.} Adapting a model during inference using signals derived from the model itself has been studied extensively for perception, typically through entropy minimization or self-consistency objectives over augmented inputs \citep{wang2020a}. Most closely related to our objective is the work of \citet{zuo2025}, who form a proxy reward from majority voting over sampled outputs and optimize it with reinforcement learning to fine-tune language models without labels. Our loss replaces the discrete vote with a variance-weighted fusion of continuous action distributions, and replaces the RL objective with direct behavioral cloning towards the consensus, which admits a single-step gradient and is therefore compatible with real-time operation.

\textbf{Online learning for recurrent networks.} RTRL \citep{williams1989} computes exact gradients for recurrent networks without storing a sequence, at a computational cost that has historically limited its use. Approximation such as RFLO \citep{murray2019} or e-prop \citep{bellec2020} reduce this cost substantially by dropping the non-local terms of the RTRL update, simultaneously removing the weight transport and separate backward phase that make BPTT biologically implausible. Recent interest in diagonal recurrent architectures has renewed attention to online gradient computation, as their structure admits efficient trace updates~\citep{zucchet2023}. We adopt RFLO as our gradient estimator, and to our knowledge this is the first use of real-time recurrent gradient computation for unsupervised recovery from distribution shift.

\section{Background}\label{background}

In this section, we review interactive imitation learning and the safety-gated DAgger variants that RTR-IIL builds on, followed by the recurrent network model and online gradient computation method our approach relies on.

\subsection{Interactive Imitation Learning}\label{interactive-imitation-learning}

In imitation learning, a novice policy is trained to predict expert actions from a sequence of observations from the environment. Historically, a dataset of expert trajectories was collected and the novice policy was subsequently trained offline. This method, known as behavioral cloning, can lead to unsatisfactory performance when deploying the novice due to overfitting on the training set. The DAgger algorithm \citep{ross2011} provides an intuitive solution to this problem: the dataset is constructed interactively by allowing the pretrained novice to act at times during task execution, with the expert providing action labels for the resulting observations. This leads to the inclusion of trajectories where the expert is forced to recover from suboptimal states. By alternating between training and collecting experience, the resulting novice policy is more robust.

Several variants restrict novice control to states where it can be trusted. SafeDAgger \citep{zhang2017a} trains an auxiliary ``doubt'' model to predict whether the novice's action would fall within a margin \(\tau_m\) of the expert's, and withholds control whenever it would not. HG-DAgger \citep{kelly2019} applies the same idea but additionally avoids interrupting the expert once they have taken over, so as not to provoke corrective actions that could worsen execution. EnsembleDAgger \citep{menda2019} replaces the doubt model with an ensemble of novice policies, gating control on both the margin criterion and an upper bound \(\tau_v\) on the ensemble's action variance -- we adopt this variant, as an ensemble also supplies the diversity our self-supervised recovery mechanism relies on (Section 3). We give the corresponding decision rule in algorithmic form in Section 3.

\subsection{RNNs and Online Gradient Computation}\label{recurrent-neural-networks-and-online-gradient-computation}

We will now introduce two instances of RNNs that allow for fast computation of online gradients and that integrate nicely into our method described below. Note, that the two models introduced shortly belong each to a separate general class of RNNs that are compatible with our approach and also allow for scaling up.

\paragraph{Continuous-Time Recurrent Neural Network (CT-RNN)}\label{a-simplified-ct-rnn-model}

A simple CT-RNN has \(N\) hidden units and \(I\) inputs, activation function \(\varphi\) and a single combined weight matrix \(W \in \mathbb R^{N\times Z}\) where \(Z=I+N+1\). Each neuron has a time-constant \(\tau \in \mathbb R^{N}\) and the next state \(x(t+1)\in \mathbb R^{N}\) is computed as follows
\[
x_{t+1} = x_{t}+\frac{1}{\tau}\left[-x_{t}+\varphi(W \xi_t)\right]
\]
where \(\xi_t = [x_t;\,u_t;\,1]^T \in \mathbb R^{Z} {}\) , \(u_{t}\) being the input at time \(t\). The output \({} \hat y_{t} \in \mathbb R^O {}\) is given by a linear mapping - i.e.~\({} \hat y_{t}=W_{out}x_t {}\). When interpreted as ODE: \(\tau\frac{dx}{dt} = -x_{t}+\varphi(W \xi_t)\).

\paragraph{Linear Recurrent Units (LRUs)}~\citep{orvieto2023} – describe their latent dynamics through a linear system, enabling highly efficient state updates. They define the following state-space model (SSM) equations:
\begin{equation}
\label{eq:lru}
\dot h(t)=A h(t) + B x(t)
\end{equation}
%
\paragraph{Online Gradient Computation}
where $A \in \mathbb{C}^{N \times N}$ is a diagonal matrix, and $B \in \mathbb{C}^{N \times I}$, $C \in \mathbb{C}^{O \times N}$, and $D \in \mathbb{R}^{O \times I}$ are learnable parameter matrices. The hidden state $h_t \in \mathbb{C}^N$ is complex-valued. The output $y(t)=\Re\left[C h(t)\right]+D  x(t)$ is obtained by combining the real part of the latent state with the input $x(t)$.
Real-Time Recurrent Learning (RTRL) \citep{williams1989} computes gradients for such a network online, in parallel with the forward pass, by maintaining a Jacobian trace \(\hat J_t\) alongside the hidden state. Writing for the input, the exact trace recursion for the recurrent weights \(W\) is
\[
\hat J_{t+1}^W = (1-\tfrac{1}{\tau})\, \hat J_t^W + \tfrac{1}{\tau}\varphi'(W\xi_t)^\top \xi_t + \tfrac{1}{\tau}\, \hat J_t^W \nabla_{h_t}\varphi(W\xi_t),
\]
where the last term propagates gradient information between units and requires \emph{weight transport} -- the same non-local dependency that makes standard backpropagation biologically implausible. This term is also what drives RTRL's high computational cost, which has limited its practical use.

Random Feedback Local Online Learning (RFLO) \citep{murray2019} is a biologically plausible approximation of RTRL that removes this obstacle in two steps: it drops the non-local term entirely, leaving a local, constant-cost trace update,
\[
\hat J_{t+1}^W \approx (1-\tfrac{1}{\tau})\, \hat J_t^W + \tfrac{1}{\tau}\varphi'(W\xi_t)^\top \xi_t,
\]
and it replaces the transposed output weights ordinarily used to propagate an error signal backward with a fixed random feedback matrix \(B\), as in feedback alignment.
The resulting parameter update, applied at every step from the error \(\varepsilon_t\) at the output, is \(\Delta\theta(t) = \eta\, B\varepsilon_t \odot \hat J_t\). \citet{murray2019} show that this update nonetheless aligns \(B\) with the readout weights over time, so that the error continues to decrease despite the approximation. We use RFLO throughout to compute the online updates for RTR-IIL and RTR-FTL described next.

\section{Methods}\label{methods}

We will now introduce Real-Time Recurrent Interactive Imitation Learning (RTR-IIL), a method that combines online gradient computation for recurrent neural networks with EnsembleDAgger to create a fully online learning algorithm that operates in a streaming fashion. It is not necessary to provide an expert label at each individual step: whenever a label is provided, the policy is updated and the novice policy takes over as soon as the expert yields control. After the expert retakes control, an immediate update that respects the recurrent influence of network parameters is possible due to the continuous updating of the Jacobian trace using RTRL/RFLO. We further propose a self-supervised follow-the-leader (FTL) approach for adding robustness to distribution shift even in the absence of an expert.

In this section, we first explain the RTR-IIL algorithm in detail followed by a presentation of the ensemble architecture that was used in our experiments, including the Kalman fusion technique that factors in the confidence of each individual ensemble module when constructing the policy's output action distribution. Finally, we present the self-supervised FTL loss function that, when optimized online, enables robustness to distribution shifts during live operation.

\subsection{Real-Time Recurrent Interactive Imitation Learning}\label{real-time-recurrent-interactive-imitation-learning}

We propose a framework for truly online interactive imitation learning that combines online gradient computation via RTRL with EnsembleDAgger to arrive at an imitation learning framework that can be used in a streaming fashion. We call our method Real-Time Recurrent Interactive Imitation Learning – or OnlineDAgger, as it is a variant of the DAgger algorithm that employs online gradient computation, as in RTRL or RFLO, to compute and apply updates immediately after the expert takes over.

Our method combines online gradient computation via RTRL and DAgger to create a fully online interactive imitation learning algorithm. We use an ensemble of RNNs to model the novice policy and train it using the gradient computed from RTRL. Specifically, in each step we also compute the expert action \(a^{*}\) and decide if the expert policy needs to be queried based on the ensemble variance and the (Euclidean) distance between expert and novice action. If the novice ensemble variance is larger than some threshold \(\tau_{v}\) or the distance if larger than some threshold \(\tau_{m}\), the policy is considered not confident enough, and policy parameters are updated in order to increase the probability of the expert actions. Particularly, if one of the two conditions is violated, the gradient of the distance between expert and novice actions with respect to the RNN's output \(\nabla_{h}||a^{*}-\hat\pi(h, \theta_{O})||\) is computed for each module of the ensemble. The resulting vectors are each combined with the respective RNN Jacobian trace \(\hat J_{t}\) to compute the updates applied to the RNN parameters. The above procedure is summarized in Algorithm \ref{alg:rtr}.

\begin{algorithm}[t]
    \caption{\small Real-Time Recurrent Interactive Imitation Learning}    
    \label{alg:rtr}
    \begin{algorithmic}
        \Require Expert policy: $\pi^{*}(a|o)$
        \Require RNN body: RNN$(o, h, \theta_{R})$
        \Require Novice policy output layer: $\hat\pi(a|o, \theta_{O})$
        \Require Decision thresholds $\tau_{ml},\ \tau_{mu},\ \tau_{vl},\ \tau_{vu}$
        
        \Function{CheckCriteria}{$s, \tau_{m}, \tau_{v}$}
            \State \Return $||\bar\pi(s) - \pi^{*}(s)||^{2} > \tau_{m}$ or $\text{Var}(\bar\pi(s))>\tau_{v}$
        \EndFunction
        
        \Function{loss}{$a_{exp}, a_{nov}$}
            \State \Return $\frac{1}{2}\sum(a_{exp} - a_{nov})^2$
        \EndFunction
        
        \State $\theta_{O}, \theta_{R} \gets $ Randomly initialize parameters
        \For{$i = 1$ to $N$}
            \State $o \gets $ Reset Environment
            \State expertControl $\gets$ False
            \For{$t = 1$ to $T$}
                \State $h, \hat J \gets $RNN$(o, h, \theta_{R})$
                \State $\bar\pi \gets FUSED(\hat\pi(h, \theta_{O}))$
                
                \State expertControl $\gets$ \Call{CheckCriteria}{$o,\tau_{mu},\tau_{vu}$}
                
                \If{expertControl}
                    \State $a^{*} \sim \pi^{*}(o)$
                    \State $\theta_{O} \gets \theta_{O} + \alpha_{O} \nabla_{\theta_{O}}$ \Call{loss}{$a^{*},\hat\pi(h, \theta_{O})$}
                    
                    \State $g \gets \nabla_{h}$ \Call{loss}{$a^{*},\hat\pi(h, \theta_{O})$}
                    \State $\theta_{R} \gets \theta_{R} + \alpha_{R} \hat Jg$ 
                    \State expertControl $\gets$ \Call{CheckCriteria}{$o,\tau_{ml},\tau_{vl}$}
                    \State $o \gets$ Execute $a \sim \mathcal N(a^{*}|\sigma^{2})$
                \Else
                    \State $o \gets$ Execute $\hat a \sim \bar\pi$
                \EndIf
            \EndFor
        \EndFor
    \end{algorithmic}
\end{algorithm}

\paragraph{Ensemble Policy}\label{ensemble-policy}

An ensemble policy is created by stacking \(N\) copies of the same network architecture, but initializing each copy differently at the beginning of training. A \emph{consensus} output is then found by combining the individual module's outputs. A naive method is to just take the mean but we will introduce a more involved method below. The variety in starting conditions leads to a diverse set of policy functions that, when combined, should generalize better than each individual policy on its own.

When using stochastic policies, each ensemble module predicts parameters of a probability distribution (e.g.~\(\mu\) and \(\sigma^{2}\) for a Gaussian) from which actions can be obtained by either taking the mode, or by sampling from the distribution. The naive consensus action distribution is attained by taking the mean of \(\mu\) and \(\sigma^2\) separately. We can also choose to employ a deterministic policy instead, in which case the individual member models directly compute continuous actions, and the output is simply their mean. Here, individual modules are trained by minimizing a differentiable distance measure such as the mean-squared-error loss.
\paragraph{Kalman Fusion}\label{kalman-fusion}

The policy of our agent is made up of an RNN ensemble where each submodel \(n\) predicts the parameters of a Gaussian distribution \({} \mathcal N(\mu_{n}, \sigma^{2}_{n}) {}\). The distributions are combined in order to get a final distribution \(\mathcal N(\bar\mu, \bar\sigma^{2})\) to sample the action from. A naive approach of combining the distributions is by taking the mean over location and variance. The issues with this approach, is that every \(\mu_n\) is weighted the same even though some members of the ensemble are more confident than others reflected by the predicted variance \(\sigma^{2}_{n}\).
Instead, we take an approach inspired by \citet{valencia2025} by employing a Kalman-gain method for sequentially fusing the policies. Specifically, we loop over the distributions produced by the ensemble and combine them using equations Eq.~\ref{eq:kalman-fusion}. For \(0<n\leq N\), we compute
\begin{equation}\protect\phantomsection\label{eq:kalman-fusion}{
\begin{aligned}
k_{n} & =\frac{\bar \sigma_{n-1}^{2}}{\left(\bar \sigma_{n-1}^{2}+\sigma_{n}^{2}\right)} \\
\bar \sigma_{n}^{2} & =(1-k_{n}) \bar \sigma_{n-1}^{2}+k_{n} \sigma_{n}^{2} \\
\bar \mu_{n} & =\bar \mu_{n-1}+k_{n}\left(\mu_{n}-\bar \mu_{n-1}\right)
\end{aligned}
}\end{equation}
where \(\bar \mu_{0} = \mu_{0}\) and \(\bar \sigma_{0}^{2} = \sigma_{0}^{2}\) are the location and standard deviation of the first ensemble module, \(k_{n}\) is the Kalman gain that weighs the confidence of module \(n\) against the intermediary fusion result \(\mathcal N(\bar\mu_{n}, \bar \sigma_{n}^{2})\), and \(\mathcal N(\bar\mu, \bar \sigma^{2}) = \mathcal N(\bar\mu_{N}, \bar \sigma_{N}^{2})\) is the final fused action distribution of the ensemble. 

Colloquially speaking, the Kalman gain is a measure of how sure each member is about their decision compared to the tentative fused distribution at each step. The mean is therefore scaled by the gain in order to reflect the module's confidence. It has to be noted that the order of fusion matters here. For this paper however, we kept the order fixed and defer investigating ordering strategies to future work.

\subsection{Recovery from Distribution Shifts via Self-Supervised Fine-Tuning}\label{recovery-from-distribution-shifts-via-self-supervised-fine-tuning}

The assumption that an expert policy is always readily available to present action labels for behavioral cloning -- although very common in the literature -- is often difficult in practice. For example, when dealing with a human expert, smooth hand-over can be tricky to accomplish and requires careful consideration. More importantly, it is often hard for a human expert to give accurate labels for vector control actions.

We show how our basic framework can be used for online fine-tuning without any expert supervision. To this end, we employ a follow-the-leader (FTL) approach through optimizing a self-supervised loss that makes use of the Kalman gain similar to the ensemble fusion method introduced above. We refer to the resulting algorithm as RTR-FTL in the following.

\paragraph{Random Observation Masking}\label{random-observation-masking}

The setting we are exploring with this experiment is a failure of a single sensor. To achieve robustness in this scenario, and in order to increase the diversity in policy functions achieved during pretraining even further, we mask the observation seen by each member of the ensemble using a randomly sampled binary mask as shown in figure \ref{fig:ensemble}. The binary mask ensures that each member learns to predict the expert actions based on a different subset of the observation dimensions.

The self-supervised objective we use for online adaptation at test time is inspired by \citet{zuo2025} who used a similar concept for fine-tuning LLMs. Their approach, called Test-Time Reinforcement Learning, makes use of majority voting to form a proxy reward function to be maximized via RL. To this end, a batch of output actions is sampled from the stochastic policy, and the consensus action is found by majority voting. Subsequently, each output is assigned a reward of $1$ if it matches the consensus answer and 0 otherwise.

\begin{table}[h]
\centering
\caption{Mean reward and standard deviation achieved for 10 seeds of each method on the three \texttt{brax} environments. }
\label{tab:from-scratch}
\begin{tabular}{
  @{}p{0.12\linewidth}
  p{0.20\linewidth}
  p{0.20\linewidth}
  p{0.20\linewidth}@{}}
\toprule
Model & ant & halfcheetah & humanoid \\
\midrule
 & RTR-IIL &\\
\midrule
rflo & 6202.12 $\pm$ 2472.04 & 4413.42 $\pm$ 715.59 & 3044.50 $\pm$ 1433.53 \\
lru\_rtrl & 6243.34 $\pm$ 190.19 & 5179.56 $\pm$ 1405.91 & 8674.39 $\pm$ 1145.41 \\
\midrule
& EnsembleDAgger & \\
\midrule
mlp & 6305.16 $\pm$ 166.11 & 5176.58 $\pm$ 912.39 & 8039.45 $\pm$ 220.53 \\
bptt & 6299.45 $\pm$ 294.41 & 5104.02 $\pm$ 125.79 & 8825.00 $\pm$ 9.45  \\
lru & 6317.83 $\pm$ 448.54 & 5072.12 $\pm$ 1947.61 & 8683.34 $\pm$ 57.17 \\
\bottomrule
\end{tabular}
\end{table}

While the method of \citet{zuo2025} makes use of self-supervised discrete labels to compute a reward that is optimized through RL, we propose to use a self-supervised signal based on the fused continuous action distribution predicted by the ensemble as a label for behavioral cloning. In particular, we compute the consensus action using the Kalman fusion method described above for combining the ensemble of Gaussian policies and use it's mean as a label for training the individual ensemble modules. The mean of the fused distribution can be considered the consensus that the ensemble modules have agreed upon. We then maximize the log probability of the consensus action under each individual ensemble module's output distribution, and scale it proportionally to the complement of the respective squared Kalman gain that results from comparing to the consensus distribution. Formally, we compute the contribution coefficient \(\bar{k}_{n}\) for each module again after the fusion and then sum up the scaled log probabilities of the fusion mean under each module's distribution
\begin{equation}\protect\phantomsection\label{eq:self-repair-loss}{
\begin{aligned}
\bar{k}_{n}&=\frac{\bar \sigma_{N}^{2}}{\left(\bar \sigma_{N}^{2}+\sigma_{n}^{2}\right)} \\
\mathcal L_{SR} &= -\sum_{n \in N} (1-\bar{k}_{n}^{2}) \log p(\bar{\mu}_{N} \mid \mu_{n}, \sigma_{n}^2)
\end{aligned}
}\end{equation}
We compute the gradient of this loss with respect to the network parameters using RFLO as described above and apply an update after each environment step. This enables the policy to immediately react to distribution shifts and effectively self-repair. Our proposed method is subsequently able to mitigate loss of control during operation.

\section{Experiments}\label{experiments}

We compare our RTR-IIL algorithm with CT-RNN and LRU base on a number of continuous control tasks from the \texttt{brax} package: \texttt{ant}, \texttt{halfcheetah} and \texttt{humanoid}, and further include as baselines existing EnsembleDAgger variants that use MLPs, or the same RNNs but trained with BPTT. We refer to the Mujoco documentation \citep{todorov2012} for a detailed description of the tasks. For the RTR-IIL locomotion experiments, we used the \texttt{spring} which is computationally cheap. For RTR-FTL, we switch to the \texttt{mjx} backend that is computationally more expensive but is less prone to exhibit illogical forces.
For both sets of experiments, we conducted a preliminary hyper-parameter sweep to determine the best set of hyper-parameters. The parameter-bounds and results of the sweep are given in the appendix.

\begin{figure*}[t]
\begin{subfigure}
{0.49\linewidth}\centering\pandocbounded{\includegraphics[ keepaspectratio,alt={Evaluation reward stays at a low value during operation without fine-tuning.}]{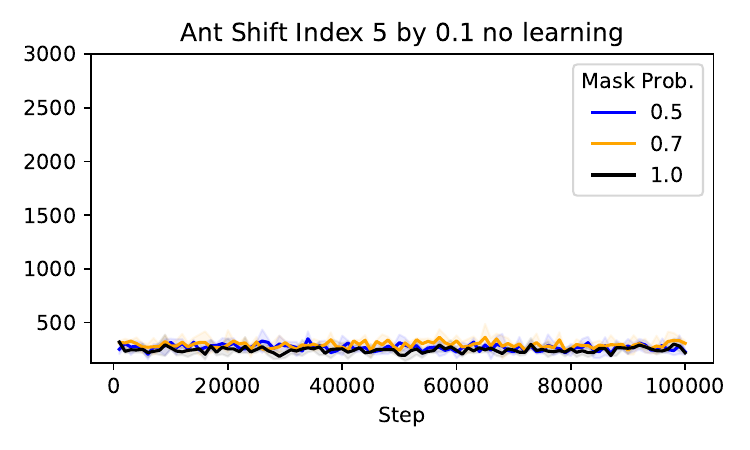}}\caption{Evaluation reward stays at a low value during operation without fine-tuning, regardless of the amount of masking.}
\end{subfigure}%
\hfill
\begin{subfigure}
{0.49\linewidth}\centering\pandocbounded{\includegraphics[ keepaspectratio,alt={While the ensemble of modules with full visibility of the entire observation is incapable of recovering, the ensemble with randomly masked observations recovers successfully. The amount of masking did not have a strong effect.}]{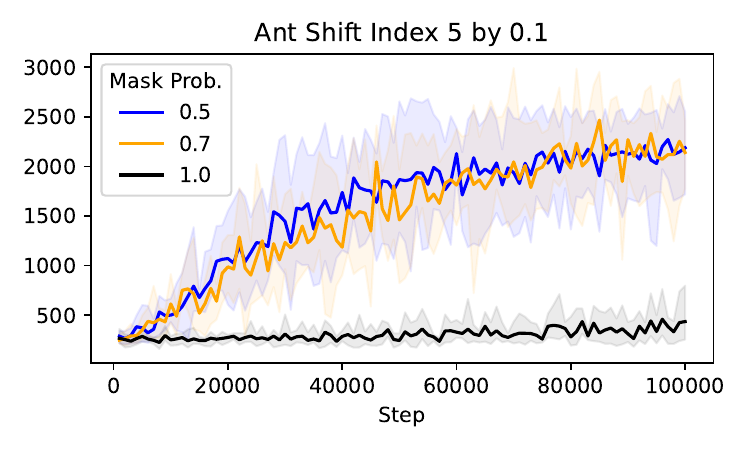}}\caption{The ensemble with randomly masked observations recovers successfully, while the ensemble of modules with full visibility does not.}
\end{subfigure}%
\caption{The pretrained controller can recover from a sudden shift in sensor reading when fine-tuned using RTR-FTL.}\label{fig:self-healing-ant-shift}
\end{figure*}

We started off by learning an expert policy using the Proximal Policy Optimization (PPO) implementation found in the \texttt{brax} package. The PPO policies serve as experts for the interactive imitation learning experiment discussed in the following sections, and trajectories collected from these policies were used for pretraining the ensembles that are later fine-tuned using RTR-FTL. We report the rewards achieved by the expert policies, and the corresponding hyper-parameters and learning curves in the appendix.

\paragraph{Training a Policy from Scratch}\label{training-a-policy-from-scratch}

In a first set of experiments, we assess the feasibility of our approach by training a novice policy from scratch. We initialize the RNN ensemble randomly and train it while acting in the environment. At each step we also query the expert action and whenever the novice action is too far from the expert action, or when the novice action variance is too high, we hand over control to the expert. We compare RTR-IIL with EnsembleDAgger variants that make use of batched replay. The results in table \ref{tab:from-scratch} show that RTR-IIL with LRU is on-par with EnsembleDAgger and performs well with CT-RNN on all environments except \texttt{humanoid} while being fully online.
\begin{figure}
    \centering   \includegraphics[width=\linewidth,alt={Scatter plot of evaluation reward vs evaluation loss for different ensemble methods and visibility masking probability.}]{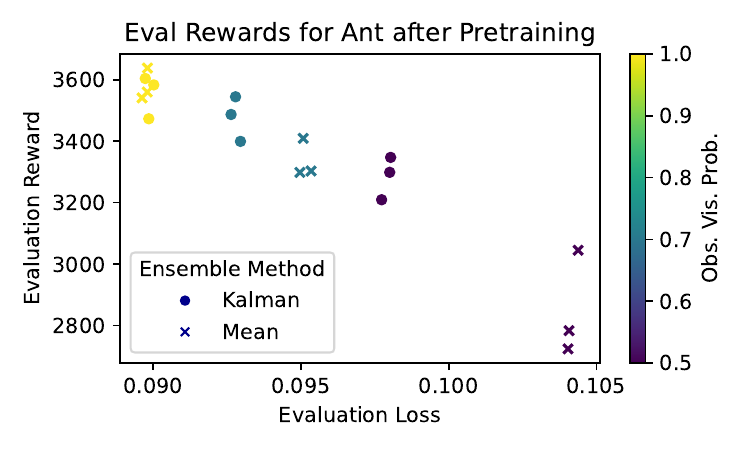}
    \caption{Scatter plot of evaluation reward vs evaluation loss for different ensemble methods and visibility masking probabilities. The Kalman fusion method retains the performance better for stronger masking.}
    \label{fig:pretrain-ensemble-method}
\end{figure}
%
%
\subsection{Self-Repair Experiments}\label{self-repair-experiments}

We now show how our masked ensemble architecture in combination with a self-supervised follow-the-leader adaptation method helps mitigating distribution shift during live operation. Our RTR-FTL method makes use of the same algorithmic framework as the previous experiments, but in place of learning from expert labels, it applies a parameter update at each step during inference, using the loss function defined in equation~\ref{eq:self-repair-loss}. In the following, we present three different types of distribution shift that our approach is able to successfully mitigate. For this, we pretrained ensemble policies from an offline dataset that contained 50 trajectories gathered from the simulated environment with the expert RL policy from earlier. Figure \ref{fig:pretrain-ensemble-method} shows a scatter plot of the evaluation loss vs. the reward after pretraining.

\paragraph{Sensor Shift}\label{sensor-shift}

Our first example is a shifted sensor reading as might occur in joints after a hard impact or after a failed homing procedure. Accordingly, we apply a shift to the observation returned by the \texttt{ant} environment at a fixed index. We evaluate three different policies that were trained with varying probability of observation masking as described above: \({} 100\%\), \({} 70\%\), and \({} 50\%\). The shift with strength \(0.1\) is applied immediately at the start of fine-tuning at position \(5\) of the observation, which corresponds to the angle between torso and the front left leg. The shift leads to a total collapse of the policies' locomotion capabilities. We apply the RTR-FTL method and fine-tune for 100 episodes with three different random seeds for each pretrained model. Additionally, we ran the same three seeds with the learning rate set to \(0\) for comparison. Fig.~\ref{fig:self-healing-ant-shift} shows how our self-repair approach is able to recover from the sensor shift. While, all the models remain unable to move without learning enabled, the policies with \(50\%\) and \(70\%\) both recover close to the original performance over the course of fine-tuning.

\begin{figure*}[ht]
\begin{subfigure}{0.49\linewidth}\centering\pandocbounded{\includegraphics[keepaspectratio,alt={As before, we see a significant decrease in collected reward after the sensor failure when fine-tuning is disabled.}]{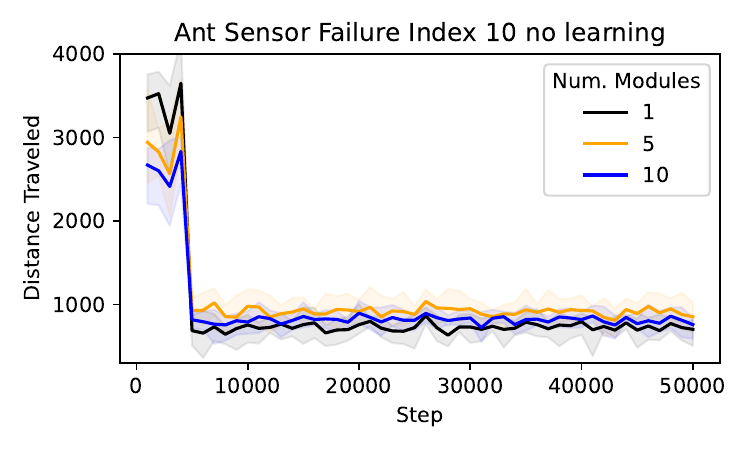}}\caption{As before, we see a significant decrease in collected reward after the sensor failure when fine-tuning is disabled.}\end{subfigure}%
\hfill
\begin{subfigure}{0.49\linewidth}\centering\pandocbounded{\includegraphics[keepaspectratio,alt={We observed that the ensemble with visibility probability of 0.5 was able to recover somewhat from the complete sensor failure. The lineplot shows that an ensemble of 5 modules performed best in this particular setting.}]{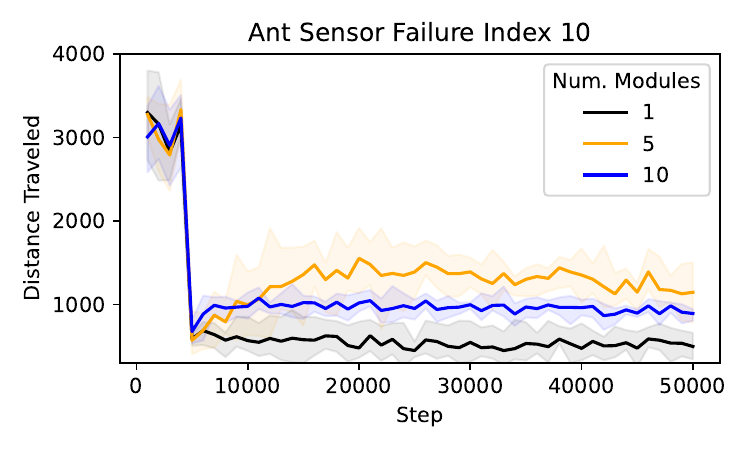}}\caption{The ensemble with visibility probability of 0.5 is able to recover to some extent from the complete sensor failure.}\end{subfigure}%

\caption{RTR-FTL can also recover from a complete sensor failure. Here, the observation at index $10$ was set to $0$ after $5k$ steps. The lineplot on the right shows that an ensemble of 5 modules performed best in this particular setting.}\label{fig:self-healing-ant-failure}
\end{figure*}

\paragraph{Sensor Failure}\label{sensor-failure}

Next, we examined the effectiveness of our proposed method in response to total sensor failure by removing a sensor reading entirely. To this end, we forced the observation of the \texttt{ant} environment at index \(10\) – which corresponds to the angle between the two links of the left back leg – to \(0\). The results in fig.~\ref{fig:self-healing-ant-failure} show a significant improvement when fine-tuning is switched on.

\begin{figure}
    \centering
\pandocbounded{\includegraphics[keepaspectratio,alt={In this setting, we again see a strong correlation between the number of modules and the distance traveled by the halfcheetah before collapse. This time, the mean ensemble method performed better which is why we are reporting results for it instead of the kalman fusion.}]{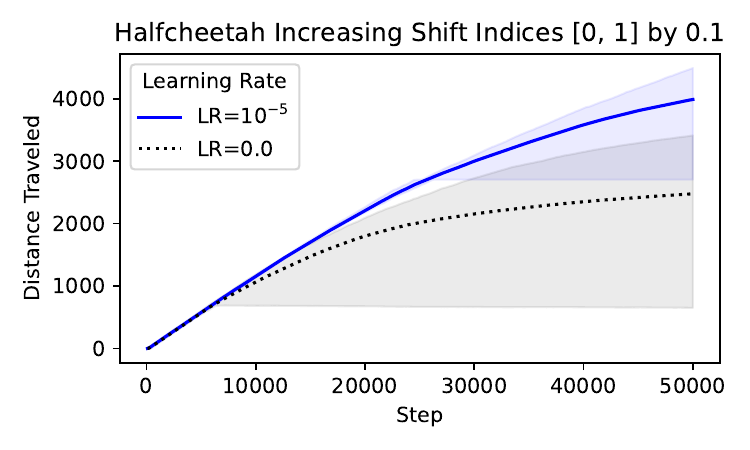}}
\caption{The line plot shows the total distance traveled by the \lstinline{halfcheetah} before collapse – without any resets of the environment – under linearly increasing shift applied to observation indices $[0, 1]$. }
\label{fig:shift-increase-distance}
\end{figure}

\paragraph{Gradually Increasing Noise}\label{gradually-increasing-noise}
Our last experiment explores the robustness to a gradual increase in sensor noise. Here, we changed the setting slightly and increased the episode length to span the entirety of the fine-tuning duration, effectively never resetting the environment. We then applied a linearly increasing Gaussian noise with $0$ mean and a maximum standard deviation of ${} 0.1$ to observation indices $[0,1]$ after $50k$ steps, and checked whether our online self-repair method could counteract the detrimental effects of the noise. The results comparing an ensemble of 10 modules with observation-visible probability of $50\%$ are presented in figure \ref{fig:shift-increase-distance}. We can see a significant improvement of the distance traveled when using our RTR-FTL method compared to the same ensemble with the learning rate set to $0$.

\section{Conclusion}\label{conclusion}

In this paper, we introduced a recurrent ensemble policy that recovers from distribution shift during deployment without expert supervision. By masking a different random subset of the observation for each ensemble member and fusing their Gaussian outputs through sequential Kalman-gain weighting, the consensus action remains reliable even when individual members are confused by a corrupted input. We use this consensus as a self-supervised label and fine-tune each member towards it in proportion to its own contribution to the fusion -- a framework we call Real-Time Recurrent Follow-The-Leader (RTR-FTL). Because gradients are computed with RFLO rather than BPTT, this update is available at every environment step, so the policy adapts to a shift as it happens rather than after the fact. The same machinery, with the consensus label replaced by an expert action, yields Real-Time Recurrent Interactive Imitation Learning (RTR-IIL), a fully online method that requires no separate training phase.

Our experiments on simulated locomotion tasks show that RTR-FTL recovers close to pre-shift performance after a sudden sensor shift, and that random observation masking is necessary for this recovery: an ensemble with full visibility of the observation collapses under the same shift and does not recover. This confirms our central claim -- that the robustness comes from the \emph{diversity} the masking induces, not from the ensemble or the fusion rule alone. We further evaluate recovery from a complete sensor failure and from gradually accumulating sensor noise, extending the sensor-shift result to failure modes that unfold instantaneously and to ones that unfold over the course of an episode.

A few limitations point to future work. It remains untested whether the approach scales to simultaneous, mixed multi-sensor corruption, or to shifts that are correlated with the agent's actions. A more realistic account of sensor unreliability seems like a natural next step, as does transferring the approach to a physical platform, where the sensor failure -- which the method is ultimately meant to address -- appears.

\bibliography{zotero}


\end{document}